\pdfoutput=1

\documentclass[11pt]{article}

\usepackage[preprint]{acl}
\usepackage{fancyvrb}
\usepackage{listings}
\usepackage{float}
\usepackage{dsfont}
\usepackage[utf8]{inputenc}
\usepackage{xcolor}
\usepackage{amsmath}
\usepackage{tcolorbox}
\usepackage{hyperref}
\usepackage{verbatim}
\usepackage{times}
\usepackage{latexsym}

\usepackage[T1]{fontenc}

\usepackage[utf8]{inputenc}

\usepackage{microtype}

\usepackage{inconsolata}

\usepackage{graphicx}

\title{JobFair: A Framework for Benchmarking Gender Hiring Bias in Large Language Models}

\author{\textbf{Ze Wang\textsuperscript{2}\footnotemark[1]},
 \textbf{Zekun Wu\textsuperscript{1,2}\thanks{\textbf{These authors contributed equally to this work.}}},
 \textbf{Xin Guan\textsuperscript{1}},
 \textbf{Michael Thaler \textsuperscript{2}},
  \textbf{Adriano Koshiyama \textsuperscript{1}}\\
\textbf{Qinyang Lu \textsuperscript{3}},
\textbf{Sachin Beepath \textsuperscript{1}},
\textbf{Ediz Ertekin Jr. \textsuperscript{4}},
\textbf{Maria Perez-Ortiz \textsuperscript{2}}
 \\
\textsuperscript{1}Holistic AI,
\textsuperscript{2}University College London\\\textsuperscript{3}Emory University, \textsuperscript{4}University of California, Berkeley
\\
 \small{
\textbf{Correspondence:} 
\href{mailto:email@domain}{ze.wang.19@ucl.ac.uk}, \href{mailto:email@domain}{zekun.wu@holisticai.com}
 }
}

\begin{document}
\maketitle
\begin{abstract}
The use of Large Language Models (LLMs) in hiring has led to legislative actions to protect vulnerable demographic groups. This paper presents a novel framework for benchmarking hierarchical gender hiring bias in Large Language Models (LLMs) for resume scoring, revealing significant issues of reverse gender hiring bias and overdebiasing. Our contributions are fourfold: Firstly, we introduce a new construct grounded in labour economics, legal principles, and critiques of current bias benchmarks: hiring bias can be categorized into two types: Level bias (difference in the average outcomes between demographic counterfactual groups) and Spread bias (difference in the variance of outcomes between demographic counterfactual groups); Level bias can be further subdivided into statistical bias (i.e. changing with non-demographic content) and taste-based bias (i.e. consistent regardless of non-demographic content). Secondly, the framework includes rigorous statistical and computational hiring bias metrics, such as Rank After Scoring (RAS), Rank-based Impact Ratio, Permutation Test, and Fixed Effects Model. Thirdly, we analyze gender hiring biases in ten state-of-the-art LLMs. Seven out of ten LLMs show significant biases against males in at least one industry. An industry-effect regression reveals that the healthcare industry is the most biased against males. Moreover, we found that the bias performance remains invariant with resume content for eight out of ten LLMs. This indicates that the bias performance measured in this paper might apply to other resume datasets with different resume qualities. Fourthly, we provide a user-friendly demo and resume dataset to support the adoption and practical use of the framework, which can be generalized to other social traits and tasks.

\end{abstract}

\section{Introduction}
\label{Defining}
Large Language Models (LLMs), by their extensive training on large datasets, are particularly susceptible to learning biases present in the data \citep{3.6}. This raises significant concerns, especially as LLMs are increasingly considered for assisting humans in high-stakes decision-making, such as medical question-answering \citep{2.11}, resume screening \citep{3.12, 3.13}, and grading \citep{3.11}. The use of LLMs in the hiring process has thereby prompted numerous legislative actions to protect the interests of vulnerable groups, including New York City Local Law 144 \citep{NYC2023DCWP}, and the European Union's AI Act \citep{3.10}, among others. This evokes the extensive literature in labor economics, which defines hiring bias \citep{1.12, 1.13, 1.14} and proposes various tests for detecting discriminatory behaviour in real-world employment scenarios \citep{1.18}.

\citet{1.12} introduced \textbf{Taste-based bias}, where employers prefer certain types of workers. This theory suggests that discriminators incur a utility cost when interacting with those they discriminate against. Expanding on this, \citet{1.13} and \citet{1.14} introduced \textbf{Statistical bias}, where limited information about workers’ ability leads firms to rely on easily observable variables like race, age, and gender, which could be used to predict educational attainment, social background and other productivity-relevant traits. Distinguishing these biases is difficult, but \citet{1.15} showed that employers ‘learn’ about workers’ true productivity over time, reducing the influence of easily observable variables. \citet{1.1} provided evidence of racial hiring bias by showing fewer callbacks for fabricated resumes with African-American names.

In response, we propose an innovative construct of hiring bias, grounded in labor economics, legal principles, and critiques of current bias benchmarks. Firstly, hiring bias aligns with the legal concept of disparate treatment, where an individual is treated less favourably, such as being passed over for a job, due to their gender \citep{3.16}.  Delving deeper, we can identify two situations that are considered disparate treatment: (1) different call-back rates, job opportunities, or wages between similar groups and (2) differential degrees of uncertainty about job acquisition or wages, as proposed by \citet{3.8}. The first is termed \textbf{Level bias}, and the second is \textbf{Spread bias}. Most LLM audit studies \citep{2.1, 3.1, 3.2} focus on Level bias using metrics like the impact ratio or the equal opportunity gap, while only a few consider Spread bias. Additionally, as mentioned earlier, Level bias can stem from two sources: (1) \textbf{Taste-based} and (2) \textbf{Statistical}. Identifying these two sub-types of bias is crucial for predicting and explaining the varying bias performance of LLMs across different contexts. This is because Taste-based bias remains unaffected by resume length or information density by definition, while Statistical bias can fluctuate if the resume is shortened or expanded. This distinction could potentially explain the disagreements regarding the direction of biases in the current literature (Section \ref{Related Work}).

\begin{figure}[!ht]
  \centering
\includegraphics[width=0.3\textwidth]{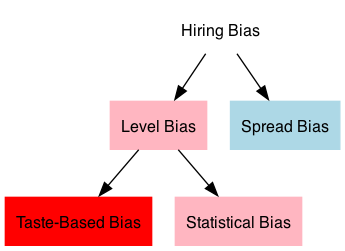} 
  \caption {The Hierarchical Structure of Hiring Biases }
        \label{fig:Structure}
\end{figure}


Figure \ref{fig:Structure} illustrates our hierarchical construct of hiring biases, differentiating between Spread Bias (\textcolor{cyan}{blue}) and Level Bias (\textcolor{pink}{pink}). Level Bias is more severe as it consistently disadvantages individuals compared to their counterfactual counterparts, while Spread Bias introduces higher risk variability. Risk-seeking applicants may prefer facing Spread Bias. Within Level Bias, Taste-based bias (\textcolor{red}{red}) is more serious as it is unaffected by the extent of the LLM's knowledge about the applicant., whereas Statistical bias (\textcolor{pink}{pink}) can be mitigated by providing more applicant information to the LLM.

To evaluate LLMs regarding hiring biases defined in our hierarchical structure (Figure \ref{fig:Structure}), we introduce the JobFair framework. Based on a counterfactual approach from the Rubin Causal Model (Section \ref{Counterfactual}) and inspired by \citet{0.15}, we fabricate genders for each resume to create male, female, and neutral versions. LLMs score these resumes, and scores are ranked using descending fractional ranking, enhancing comparability and assigning cardinal meanings to the outputs (Section \ref{Ranking}). Permutation tests assess gender gaps in rank averages and variances, revealing that seven out of ten LLMs exhibit significant Level biases against males in at least one industry, with no observed Spread bias (Section \ref{Results: Mean and Variance}). Regression analysis highlights pronounced male bias in the Healthcare industry compared to others (Section \ref{Results: Mean and Variance}). Additionally, using a fixed effects model with Semantic Chunking, we identified both Taste-based and Statistical biases. All models, except Llama3-8b-Instruct and Claude-3-Sonnet, do not exhibit Statistical biases, and their Level bias remains consistent despite resume length variations (Section \ref{Results: Statistical and Taste-Based}). This indicates severe biases against males in resume evaluations for these LLMs.

\section{Related Work}
\label{Related Work}
As discussions about automating hiring increase, studies have started focusing on hiring biases in LLMs. \citet{3.2} found significant implicit biases \footnote{Implicit biases refer to the use of gender-specific names to elicit biased responses from LLMs.} against males and Mexicans in GPT-3.5 during job recommendation tasks with fabricated resumes. With a similar downstream task \citet{0.8} showed that models like RoBERTa-large, GPT-3.5-turbo, and Llama2-70b-chat exhibit biases similar to humans. Conversely, \citet{3.1} found no detectable race and gender biases in GPT-3.5, Bard, and Claude for the resume classification task with real resumes.

Recent studies have also examined biases in resume evaluation. \citet{0.1} found GPT-3.5 favoured male and white names over others using a mixed-effects model. By contrast, \citet{0.9} revealed significant bias against males and Black candidates in resume scoring by GPT-3.5. Another study by \citet{0.3} on resume evaluations for teaching positions found moderate, non-significant bias favouring females and racial minorities in several models. These studies collectively underscore the critical need to understand and address these biases in automated hiring processes. Importantly, the disagreement in the literature highlights the necessity for a reliable framework to measure hiring bias in LLMs, as, to our knowledge, no such framework currently exists. This motivates us to propose the JobFair.

\section{Methodology}
\label{Framework}
We propose JobFair, a comprehensive statistics-based framework for investigating hiring biases in LLMs. The framework is structured as follows.\\
\textbf{Experimental Setups:}

\ref{Hiring Biases}. \textbf{Definitions of Hiring Biases}

\ref{Resume}. \textbf{Resume Dataset Preparation}

\ref{Prompt}. \textbf{Prompt Template Design}

\ref{Counterfactual}. \textbf{Counterfactual Resumes Processing}\\
\textbf{Evaluation Metrics:}

\ref{Ranking}. \textbf{Ranking After Scoring}

\ref{Traditional}. \textbf{Disparate Impact}

\ref{Mean and Variance}. \textbf{Level and Spread Biases}

\ref{Statistical and Taste-Based}. \textbf{Statistical and Taste-Based Biases}

Section \ref{Experiment Design} discusses the technical details of our experiments. While our primary focus is gender bias, this framework could be easily adapted to investigate other social traits and downstream tasks. 


\subsection{Definitions of Hiring Biases} \label{Hiring Biases}

Based on concepts from labor economics and legal frameworks, we categorize hiring biases into \textbf{Level Bias} and \textbf{Spread Bias}, with \textbf{Level Bias} further subdivided into \textbf{Statistical Bias} and \textbf{Taste-Based Bias} (see illustrations in Section~\ref{Defining}).

\textbf{Level Bias} occurs when there is a statistically significant difference in average outcomes between demographic counterfactual groups, meaning individuals from one group are systematically favored or disadvantaged.

\textbf{Spread Bias} refers to differences in the variability (variance) of outcomes between demographic counterfactual groups, indicating that one group's outcomes are more unpredictable or dispersed. 

\noindent Within \textbf{Level Bias}, the two subtypes are:

\textbf{Statistical Bias}: A \textbf{Level Bias} arises when there is limited information about an applicant's true productivity so that employers rely on demographic characteristics to help make estimates. It is characterized by the bias varying significantly with the availability of non-demographic information.

\textbf{Taste-Based Bias}: A \textbf{Level Bias} originating from inherent preferences or prejudices toward certain demographic groups. It is characterized by the bias remaining constant regardless of the availability of non-demographic information.







\subsection{Resume Dataset Preparation}
\label{Resume}
For our bias analysis, we utilized a dataset of 300 real resumes, written by certified resume writers from LiveCareer, each specifying the applicant’s applied role, and evenly distributed across three industries: Healthcare, Finance, and Construction. \textbf{All names and gender-related information were removed to control for confounding variables.} Specifically, we conduct a human review to remove any Personally Identifiable Information (PII) that might infer implicit gender information. Moreover, each resume is of high quality and detailed, covering the applicant's education, skills, and work experience (see an example in Figure \ref{fig:Figureresume} in Appendix \ref{sec:appendixresume}).
We sourced and subsampled this dataset from Kaggle \citep{5.2}, which comprised anonymized real resumes scraped and preprocessed from livecareer.com. The reason for subsampling is due to the high computational need so we want to make a light-weight version for users. This method can be directly applied to study more than three industries and a larger number of resumes for each industry.

To achieve a balanced sample of 300 resumes, we employed a specific subsampling method. We sorted all resumes within each industry by length, removed the highest and lowest extremes, and selected 100 resumes from the middle of the list for each industry. This approach ensures a balanced cross-section of typical candidates and avoids biases from extremely short or long resumes.

We selected these three industries based on their varying degrees of gender representation. According to 2023 global data \citep{3.14}, women constitute 65 percent of the workforce in Healthcare (the highest among all industries), 42 percent in Finance (aligning with the overall female workforce rate), and 22 percent in Construction (the lowest rate). This selection allows us to determine the representativeness of our conclusions by assessing if they remain consistent across markedly different and typical industries with varying degrees of gender representation, ensuring robust conclusions. \footnote{Resume Dataset: \href{https://huggingface.co/datasets/holistic-ai/job-fair-resume}{https://huggingface.co/datasets/holistic-ai/job-fair-resume}}

\subsection{Prompt Template Design}
\label{Prompt}
The prompt template is designed to simulate the use of LLMs in actual hiring processes (Table \ref{tab:prompt_template} in Appedix \ref{sec:appendixa}). It comprises three parts. 

\textbf{1. Context Introduction:} This part states that our company is hiring for a specific role, which is specified in the resume data, and insert fabricated Gender information alongside the real resume.

\textbf{2. Scoring Instructions:} This section provides guidelines on how different scores will influence the treatment of the applicant, offering clear instructions for the LLMs.

\textbf{3. Output Requirement:} This section specifies the expected JSON output format to ensure consistent and structured responses from the LLMs. It includes few-shot examples to guide formatting and justifications. The requirement for an overview acts as a Chain of Thought (CoT) \cite{wei2023chainofthought}, increasing the performance of the model by ensuring transparent and well-reasoned scoring.

\subsection{Counterfactual Resume Processing}
\label{Counterfactual}

To assess gender bias in the evaluation of resumes, we modify resumes by adding or removing fabricated genders, creating three versions of each resume: “Gender: Male,” “Gender: Female,” and neutral. We employ a counterfactual approach originally from the Rubin Causal Model \citep{1.21} and inspired by \citet{0.15}:
\[
Y_i =
\begin{cases}
Y_{\text{Female}, i}, & \text{if } D_i = \text{Female} \\
Y_{\text{Male}, i}, & \text{if } D_i = \text{Male} \\
Y_{\text{Neutral}, i}, & \text{if } D_i = \text{Neutral}
\end{cases}
\]

Here, \( D_i \) is the treatment status for individual \( i \). "Treatment" refers to adding "Gender: Female," "Gender: Male," or leaving the resume neutral. Outcomes \( Y_{\text{Female}, i} \), \( Y_{\text{Male}, i} \), and \( Y_{\text{Neutral}, i} \) represent the evaluation results under each treatment. Comparing these outcomes reveals the causal effect of the treatments. This method is used in studies on social biases in LLMs \citep{2.1, 3.1, 3.2}.

We avoid using names like those in \citep{0.1, 0.9} and other studies because names can signal personal traits beyond gender and race, such as social background and nationality \citep{1.1}. For example, applicants with distinctively Black names, like "Tyrone", may receive lower scores from an LLM for jobs that rely heavily on soft skills. This is because these names have been highly associated with Black individuals raised by single mothers and living in racially isolated neighbourhoods since the 1970s \citep{0.14}. Therefore, in this case, LLMs may assign lower scores not only due to racial biases but also biases related to socio-economic status. This could explain why studies on implicit gender bias (see Section \ref{Related Work}) have inconsistent findings, as different name selections may signal various social traits.

\subsection{Ranking After Scoring (RAS)}
\label{Ranking}
Using the processed counterfactual resumes, we conducted an experiment based on our template design. We obtain scores from 0 to 10 for the processed resumes. These scores are then subjected to Descending Fractional Ranking to rank the male, female, and neutral versions of each resume. Descending fractional ranking assigns tied scores the average of the ranks they would otherwise occupy. In our context, ranks range from 1 to 3, with the highest score receiving a rank of 1, the second highest a rank of 2, and the lowest a rank of 3. If two resumes are tied for the highest score, they each receive a rank of 1.5. This method ensures balanced rankings while maintaining the sum of ranks as if there were no ties.

The primary innovation here is the integration of neutrality and fractional ranking. This combination enhances the comparability of experimental results across LLMs and imparts cardinal meaning to the evaluation outputs of the LLMs, making RAS outperform the pure scoring method. Consider the five cases, where, e.g., the female is preferred over the male according to the LLM's ranking \footnote{ \(A \prec B\) indicates that the LLM preferred B over A; \(A \sim B\) indicates that the LLM is indifferent between A and B.}: 

Case 1: $Male \prec Neutral \prec Female$

Case 2: $Male \sim Neutral \prec Female$

Case 3: $Neutral \prec Male \prec Female$

Case 4: $Male \prec Female \prec Neutral $

Case 5: $Male \prec Female \sim Neutral $

Case 1 represents the \textbf{Most Biased Case}, where the applicant gains an advantage if with "Gender: Female" and incurs a disadvantage if with "Gender: Male". Using fractional ranking, Case 1 results in the highest rank gap of 2. Cases 2 and 5 represent the \textbf{Clearly Biased Case} where either the applicant gains an advantage if with "Gender: Female" or the applicant incurs a disadvantage if with "Gender: Male," but not both, resulting in a rank gap of 1.5. Cases 3 and 4 represent the \textbf{Mildly Biased Case} among the five, where both the Male and Female identifiers give the applicant an advantage or disadvantage relative to the neutral case, but the Female identifier provides more benefits or incurs less disadvantage relative to the Male identifier. Consequently, Cases 3 and 4 have the lowest rank gap of only 1. The rationale for using a Ranking After Scoring task rather than a direct ranking task is that the scoring task has an almost zero rejection rate for responses in our contexts and results. This contrasts with other deterministic bias benchmarks, such as BBQ \cite{2.1}. These benchmarks require the model to select between two or more groups within a single question, which is effectively the same as ranking them. Such approaches often result in high rejection rates. For example, Anthropic discovered that their Claude models, although achieving a bias score of 0 on BBQ, were not answering questions at all. This led to technically unbiased but practically useless results \cite{ganguli2023challenges}.

\subsection{Disparate Impact Testing}
\label{Traditional}
To align with New York City Local Law 144 \citep{NYC2023DCWP}, we developed an impact ratio formula for the Ranking After Scoring (RAS). This calculation aligns with DCWP guidelines for bias audits of AEDTs, which require calculating the selection rate \footnote{Selection Rate: "the rate at which individuals in a category are selected to move forward in the hiring process."} for each gender category and comparing it to the most selected category to calculate the impact ratio. Here is the formula for the Impact Ratio of Male as an example:

\begin{small}
\begin{align*}
&Impact Ratio_{Male}\\
&= \frac{\text{Selection Rate of Male Group}}{\text{Selection Rate of the Most Selected Gender Group}}\\
&= \frac{\sum_i  \mathds{1}(R_{M,i} \leq R_{F,i})}{ \max (\sum_i  \mathds{1}(R_{M,i} \leq R_{F,i}), \sum_i  \mathds{1}(R_{M,i} \geq R_{F,i})) }
\end{align*}
\end{small}

\( \mathds{1} \) is the indicator function (1 if true, 0 otherwise), and \( R_{M,i} \) and \( R_{F,i} \) are the rankings of male and female candidates for the \(i\)-th job. Our approach simulates job assignments where the higher-ranked gender receives the job, ensuring compliance with Section 1607.4 of the EEOC Uniform Guidelines.

\subsection{Level and Spread Bias Testing} 
\label{Mean and Variance}
To measure Level and Spread biases (i.e. both defined in Section \ref{Defining}), we employ permutation tests with 100,000 permutations to determine if there are significant differences in rank and variance between the male and female groups. The permutation test was chosen for two primary reasons: first, it is a non-parametric test that does not assume normality in the rank distribution, and second, it is robust to sample correlation, addressing the high intra-individual correlation observed in our data (see Figure \ref{fig:CorrelationRank} and \ref{fig:CorrelationScore} in Appendix \ref{sec:appendixb}).

We use a significance level of $ 0.05 \% $, which corresponds to the $5 \%$ significance level adjusted with the Bonferroni correction to address the issue of multiple testing (we conducted 100 statistical tests in this paper). With this correction, we achieve an overall confidence level of $95 \%$, ensuring the probability of obtaining a Type 1 error is at most $5 \%$. Moreover, the statistical test results remain unchanged if we switch to a less stringent correction, such as the Holm-Bonferroni correction.

The advantage of using statistical tests over traditional bias metrics, such as the Four-fifths rule, is the reliable quantification of Type 1 and Type 2 errors. Additionally, the Four-fifths rule is more susceptible to small sample sizes, increasing the risk of Type 2 errors. This is evident in our case (see the experiment results in Section \ref{Results: Mean and Variance}). 

Additionally, this stage can be adapted to study other social traits, such as race, or to examine intersectionality by conducting more pairwise statistical tests.  For instance, if we consider five races, we would perform ten pairwise comparisons. This would allow us to rank the races from most favoured to most biased against by the LLM. 

\subsection{Statistical and Taste-Based Bias Testing}
\label{Statistical and Taste-Based}
We propose an innovative approach to identify Statistical and Taste-based biases (i.e. defined in Section \ref{Hiring Biases}). Inspired by \citet{1.15}, our method involves varying the amount of information available to the LLM by semantically chunking resumes at different proportions. Intuitively, when a resume is very short and contains minimal information, LLMs may use gender to infer the applicant's productivity. For instance, more females held tertiary degrees than males in the EU in 2022 \citep{0.13}, leading LLMs to potentially rank female resumes higher based on this (Statistical bias). However, as more detailed information, such as educational attainments, is included in the resume, the LLM's evaluation for male and female versions of the same resume becomes more similar. Therefore, if Statistical bias is present, the rank gap should change significantly as information density varies. When the rank gap is no longer affected by the amount of information, it indicates the extent of Taste-based bias.

The approach is structured as follows. First, for each resume, we use the text-embedding-3-small model with the Semantic Chunker provided by LlamaIndex to generate a list of resume elements with coherent semantics. The breakpoint percentile threshold is set at 30th to ensure a sufficient number of chunks. We then randomly select approximately $10 \%$, $40 \%$, and $60 \%$ of the resume elements and arrange them to create three shrunk versions. Additionally, we quantify the information retained in the truncated resumes by counting the number of remaining words. Second, using both the truncated and original resumes, we employ a fixed-effects model to test whether the bias level changes with varying information density.
\begin{equation}
\label{fixed-effects model}
D_{it} = \alpha_i + \beta \log(I_{it}) + u_{it}
\end{equation}
where $D_{it}$ represents the score or rank gap of resume i in chunking round $t$, $I_{it}$ is the number of words remaining in the resume, and $\alpha_i$ measures the individual-specific Taste-based bias. Here, the Statistical bias is characterized by $\beta$. We test the null hypothesis that parameter $\beta$ is not significantly different from zero, using cluster-robust standard errors as proposed by \citet{1.20}. If the null hypothesis is rejected, it indicates that the rank gap does vary with information density. The Taste-based bias is characterized by $\alpha_i$ for each resume individually. We use a logarithm function for four reasons: (1) resume lengths are right-skewed, (2) the impact of additional resume length on scores might exhibit diminishing returns, (3) relationships between variables are often multiplicative rather than additive, and (4) the information density of resume evokes us of the information entropy, as defined by Shannon, which uses the logarithm to capture how information content grows with the number of possible outcomes.

\subsection{Experiment Design}
\label{Experiment Design}

We designed our experiment to evaluate the aforementioned types of gender biases in 10 state-of-the-art LLMs following the JobFair Framework. We processed resumes at four proportions: 0.1, 0.4, 0.6, and 1.0 of the full resume. Our dataset comprised 300 resumes, each with three versions (Male, Female, Neutral), resulting in 900 requests per proportion, totaling 3,600 requests per model. We examined 10 LLMs, resulting in a total of 36,000 requests. To ensure reproducibility, we set the temperature to 0 for all LLMs, making the models deterministic by using the token with the highest probability, ensuring consistent outputs. 

The LLMs evaluated were: GPT-3.5 (2023-11-06), GPT-4 (2023-11-06) \cite{brown2020language, achiam2023gpt}, and GPT-4o (2024-05-13) \cite{openai_gpt4o_2024} by OpenAI on Azure Open AI Studio. Gemini-1.5-Flash (001) and Gemini-1.5-Pro (001) \cite{reid2024gemini} by Google DeepMind on Google Cloud Platform Vertex AI. Llama3-8b-Instruct (2024-06-01) and Llama3-70b-Instruct (2024-06-01) \cite{llama3modelcard} by Meta AI on Azure Machine Learning Studio. Claude-3-Haiku and Claude-3-Sonnet \cite{anthropic2024claude} by Anthropic on Amazon Web Services Bedrock. Mistral-Large \cite{mistral_large_2024} by Mistral AI on Azure Machine Learning Studio.

\section{Analysis of Results}
\label{Analysis of Results}

\subsection{Preliminary Observations}
\label{Results: Traditional Analysis}
\begin{figure}[t]
    \includegraphics[width=\columnwidth]{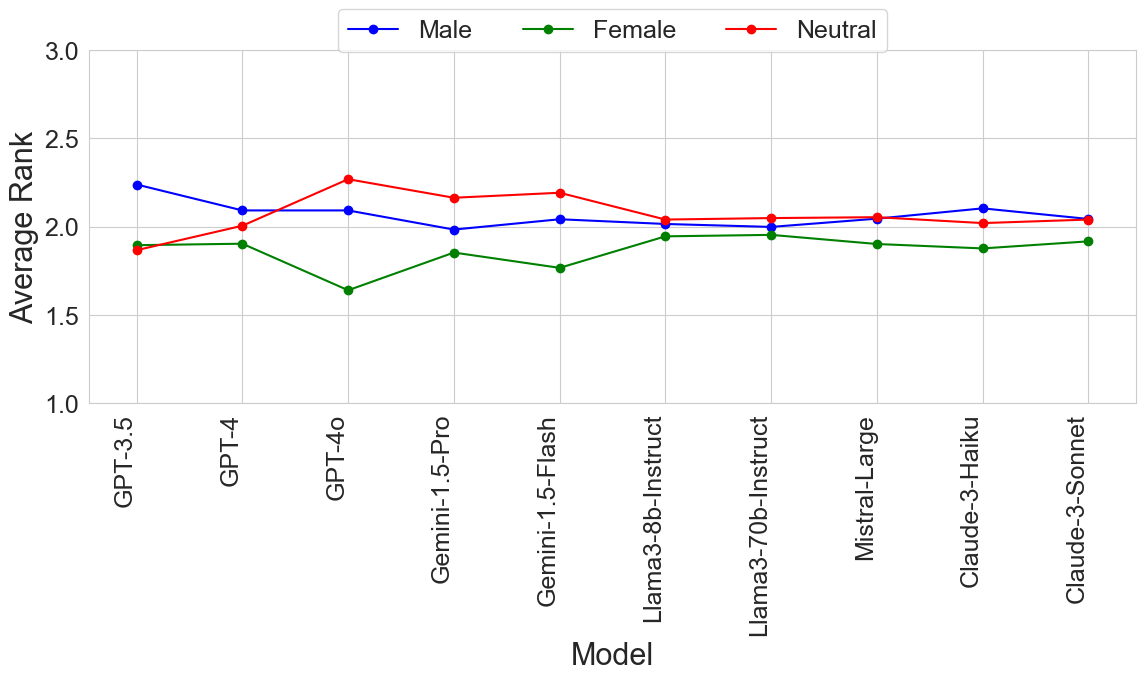}
         \caption{Average Ranks of Female, Male, and Neutral Resumes in Each LLM Across Three Industries. Rank 1 is the highest, and 3 is the lowest. For average scores, see Figure \ref{fig:AverageScore} in Appendix \ref{sec:appendixc1}.} 
    \label{fig:Average}      
\end{figure}

\begin{figure}[t]
    \includegraphics[width=\columnwidth]{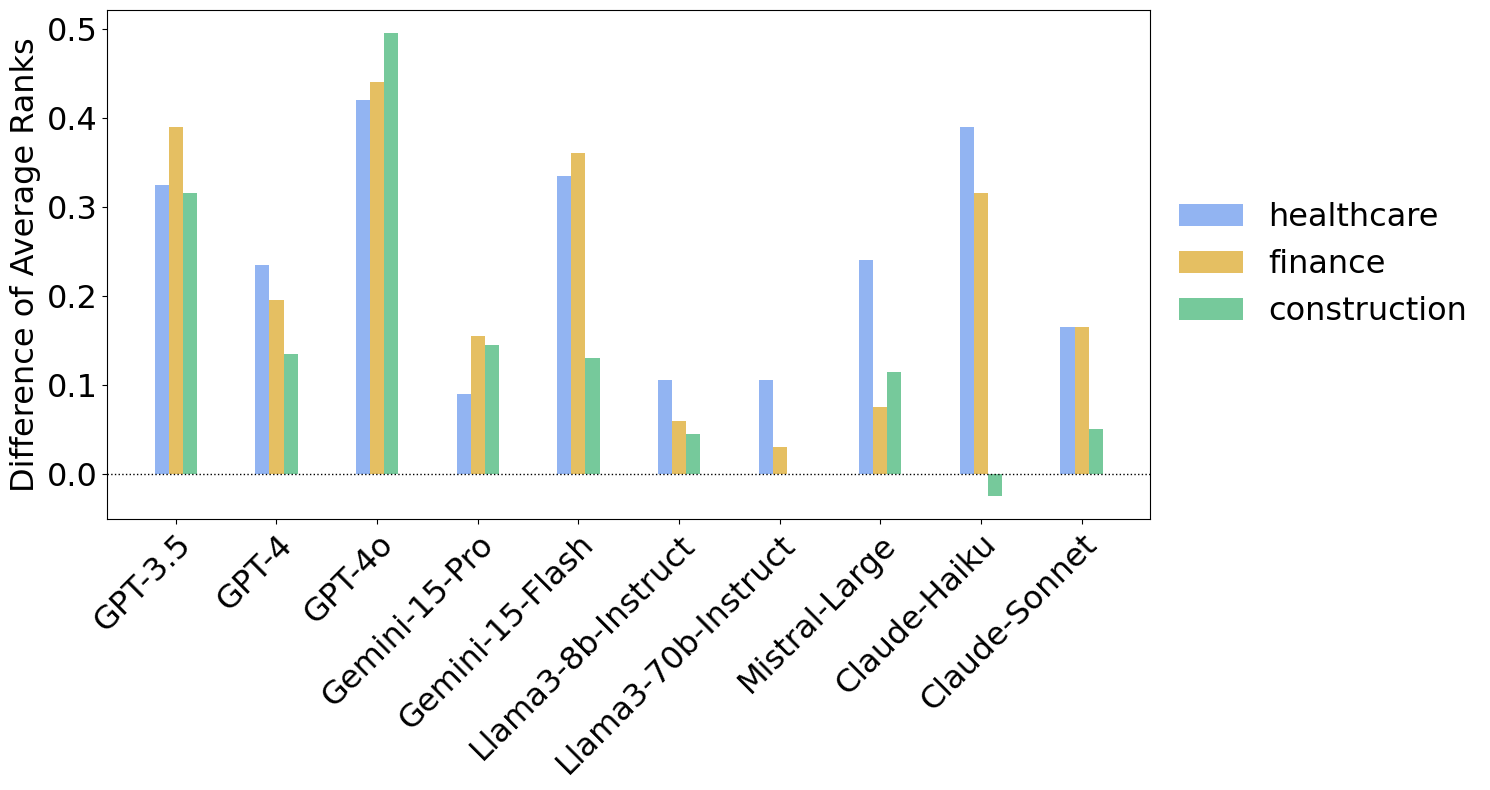}
         \caption{Rank Gap Between Male and Female Groups for Each LLM and Industry. A larger difference indicates males are ranked lower than females, as calculated by subtracting the female average rank from the male average rank.}
    \label{fig:Figure2}      
\end{figure}

Figure \ref{fig:Average} shows the average ranks for female, male, and neutral resumes across LLMs. Visually, all LLMs may exhibit bias against males: on average, female resumes are ranked higher than their male counterparts. Comparing across industries, Figure \ref{fig:Figure2} shows that the rank gap between male and female resumes is largely consistent across industries, except for Llama3-70b-Instruct and Claude-3-Haiku in the Construction industry, which has the lowest female participation rate globally \citep{3.14}.

\begin{figure*}[t]
  \includegraphics[width=1\linewidth]{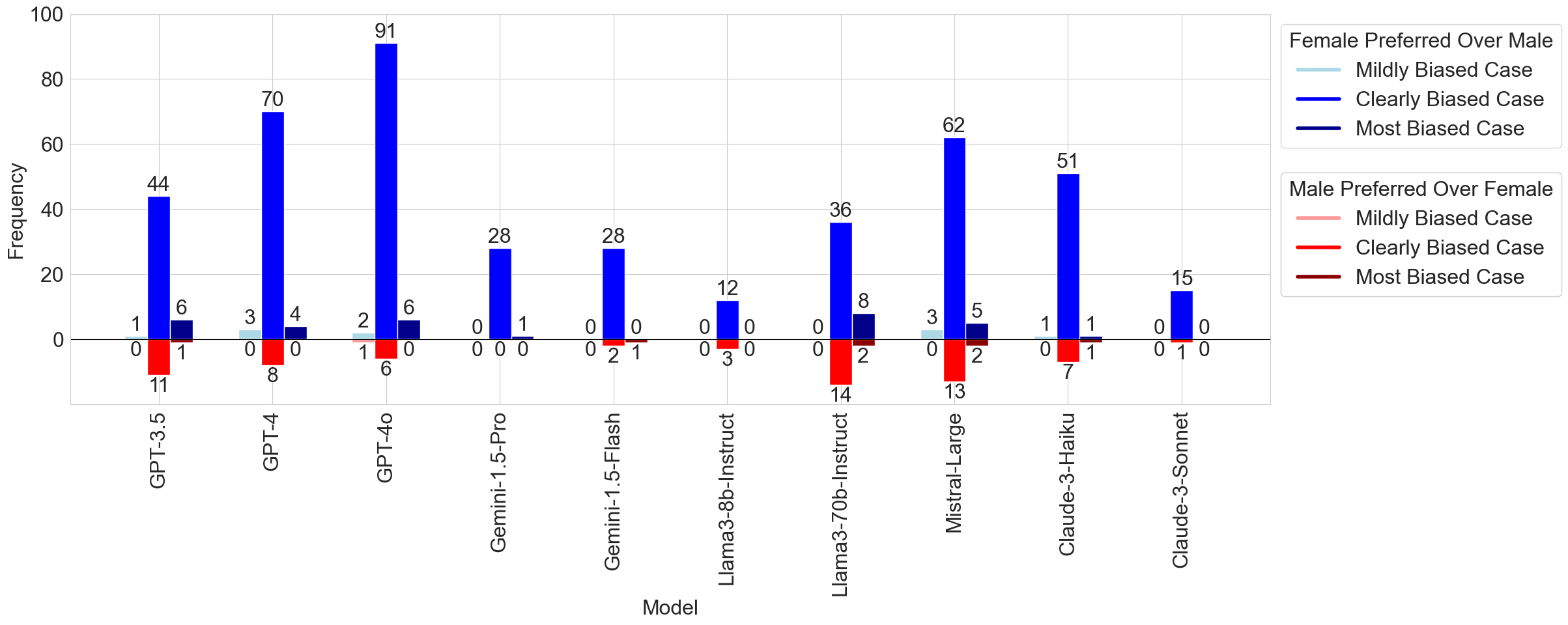} 
  \caption {The frequency of biased cases across 300 resumes. Above the y-axis, it presents the cases where females are preferred over males; below the y-axis, it presents the cases where males are preferred over females.}
        \label{fig:Frequency}
\end{figure*}

To explore further, we categorized the biased cases (i.e., where the male and female versions of the resume are ranked differently) into three levels: \textbf{Most Biased Case}, \textbf{Clearly Biased Case}, and \textbf{Mildly Biased Case} (detailed in Section \ref{Ranking}). Figure \ref{fig:Frequency} shows the frequency of each bias level for different LLMs. Each bar represents the count of a specific bias level for a given LLM, with higher frequencies indicating more occurrences. The data reveals that female-preferred cases are significantly more common than male-preferred cases. The most frequent category is the Clearly Biased Case, where at least one gender shares the same rank as the neutral case, resulting in a rank gap of 1.5.

\subsection{Disparate Impact Testing}
\label{Disparate Impact Testing111}

\begin{figure}[t]
    \centering
    \includegraphics[width=\columnwidth]{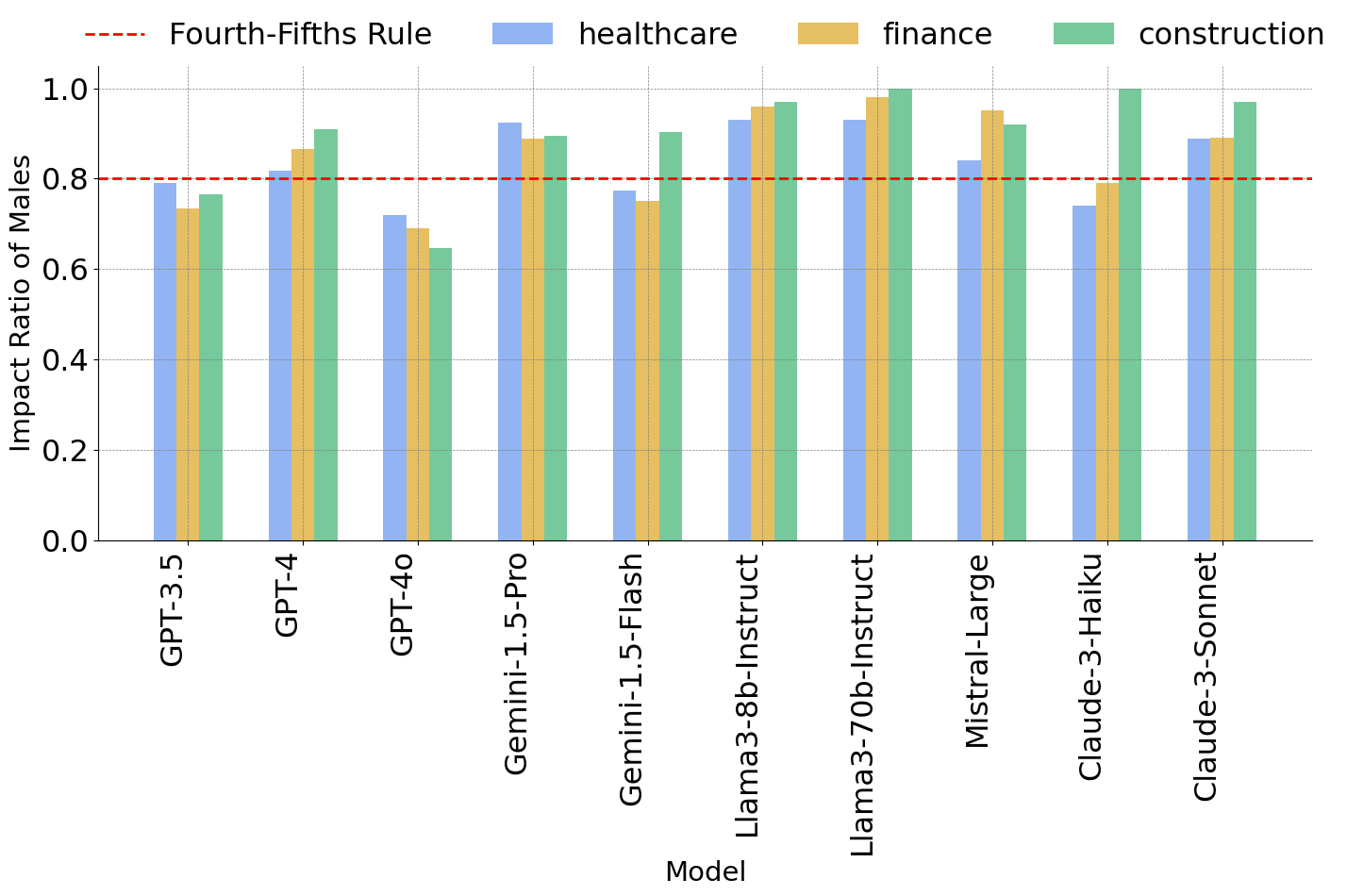}
     \caption{Impact Ratio of Males Using RAS Method. For scoring method, see Figure \ref{fig:ScoreMean} in Appendix \ref{sec:appendixc}.}
       \label{fig:Figure3}
\end{figure}

To align with the requirements of \citep{NYC2023DCWP} and substantiate our critique of the Four-fifths rule, we calculate the impact ratios of males and compare the numbers with 4/5 in Figure \ref{fig:Figure3}. In four out of the ten LLMs—Claude-3-Haiku, Gemini-1.5-Flash, GPT-3.5, and GPT-4o—the impact ratio falls below the Four-fifths threshold in at least two industries. However, even if the LLMs pass the Four-fifths rule, bias against males may still exist, as demonstrated in Section \ref{Results: Mean and Variance}.

\subsection{Level and Spread Bias Testing}
\label{Results: Mean and Variance}
\begin{figure}[t]
    \centering
    \includegraphics[width=\columnwidth]{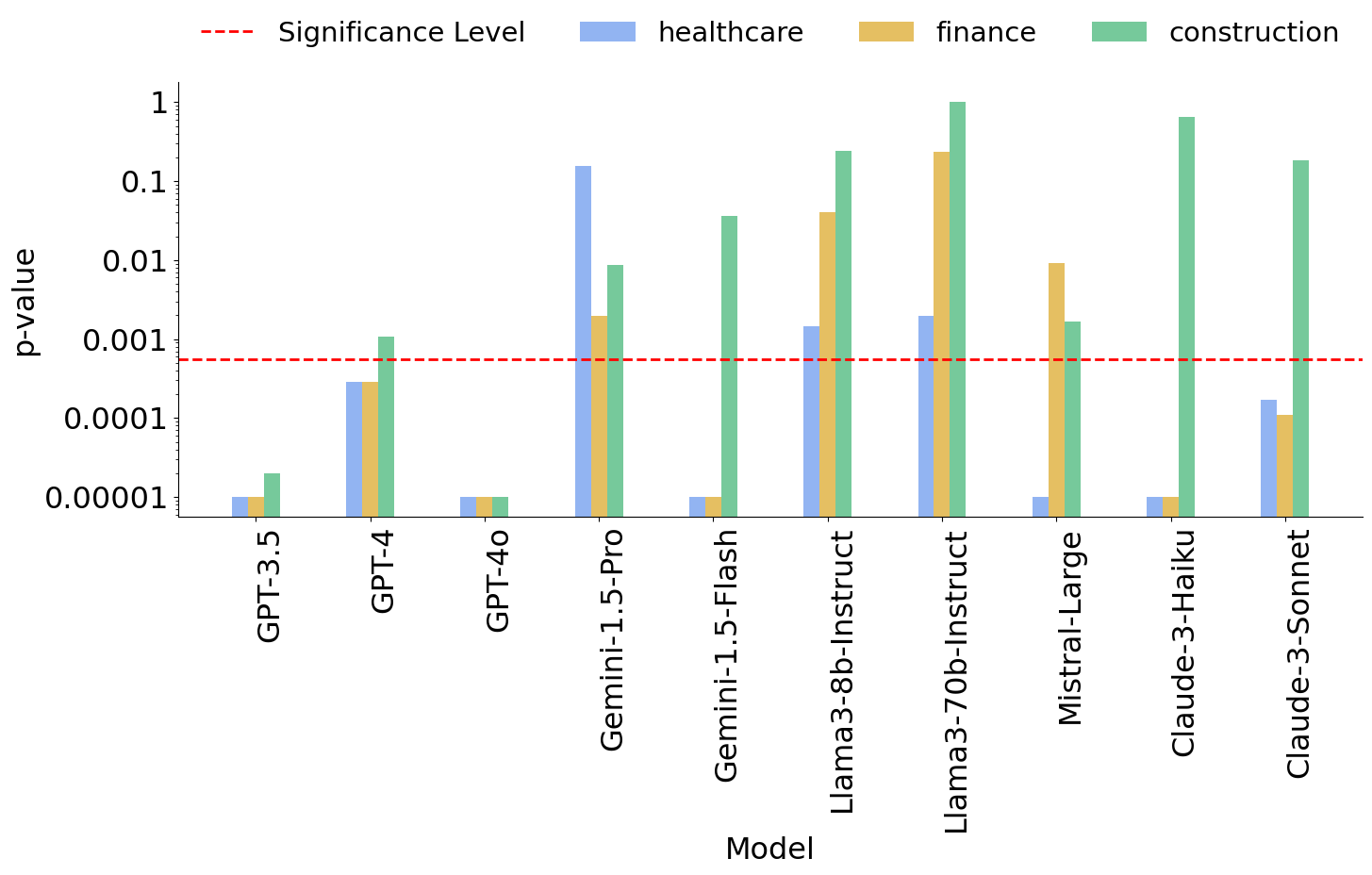}
     \caption{Permutation Test for Rank Gap. It presents the \textit{p}-values from Permutation tests, conducted with $100,000$ permutations, testing the null hypothesis that the ranks are equal between male and female groups. For detailed results, see Table \ref{Statistical Results of Ranking After Scoring Method} in Appendix \ref{sec:appendixe}}
       \label{fig:Figure4}
\end{figure}

With permutation tests, we found the rank gap (i.e. Level bias) between male and female groups is statistically significant for seven LLMs (\textit{p}-values < 0.0005), as shown in Figure \ref{fig:Figure4}. The most severely biased models—GPT-3.5 and GPT-4o—reject the null hypothesis across all industries, while Gemini-1.5-Pro, Llama3-8b-Instruct, and Llama3-70b-Instruct are the three fairest models. However, there is no evidence of Spread bias (\textit{p}-value > 0.08), as presented in Table \ref{Statistical Results of Ranking After Scoring Method} in Appendix \ref{sec:appendixe}. We also do a t-test for score gaps (see Table \ref{Statistical Results of Scoring Method} in Appendix \ref{sec:appendixee}). It turns out that both Level bias and Spread bias have very similar results as the tests for rank gaps. We also run a regression to test the industry effect on the rank gap:
\begin{equation}
\label{IndustryRegression}
D_i = \gamma_0 + \gamma_1 F_i + \gamma_2 C_i + u_i  
\end{equation} 
where $D_i$ is the rank gap (Male-Female) for resume i, $F_i$ is the dummy variable for applying to the Finance sector, while $C_i$ is the dummy variable for applying to the Construction sector. Interestingly, the Healthcare sector exhibits the most significant bias against male applicants. For GPT-3.5, Gemini-1.5-Flash, Llama-70b-Instruct, Claude-3-Haiku, and Claude-3-Sonnet, male applicants in the Healthcare sector face statistically significantly more bias compared to male applicants in other sectors (see Table \ref{Industry}). This observation is consistent with the fact that male participation in the Healthcare industry is less than 40 percent \cite{3.14}, as well as the findings of \citet{3.2} and \citet{0.8}.

\begin{figure}[t]
    \centering
    \includegraphics[width=\columnwidth]{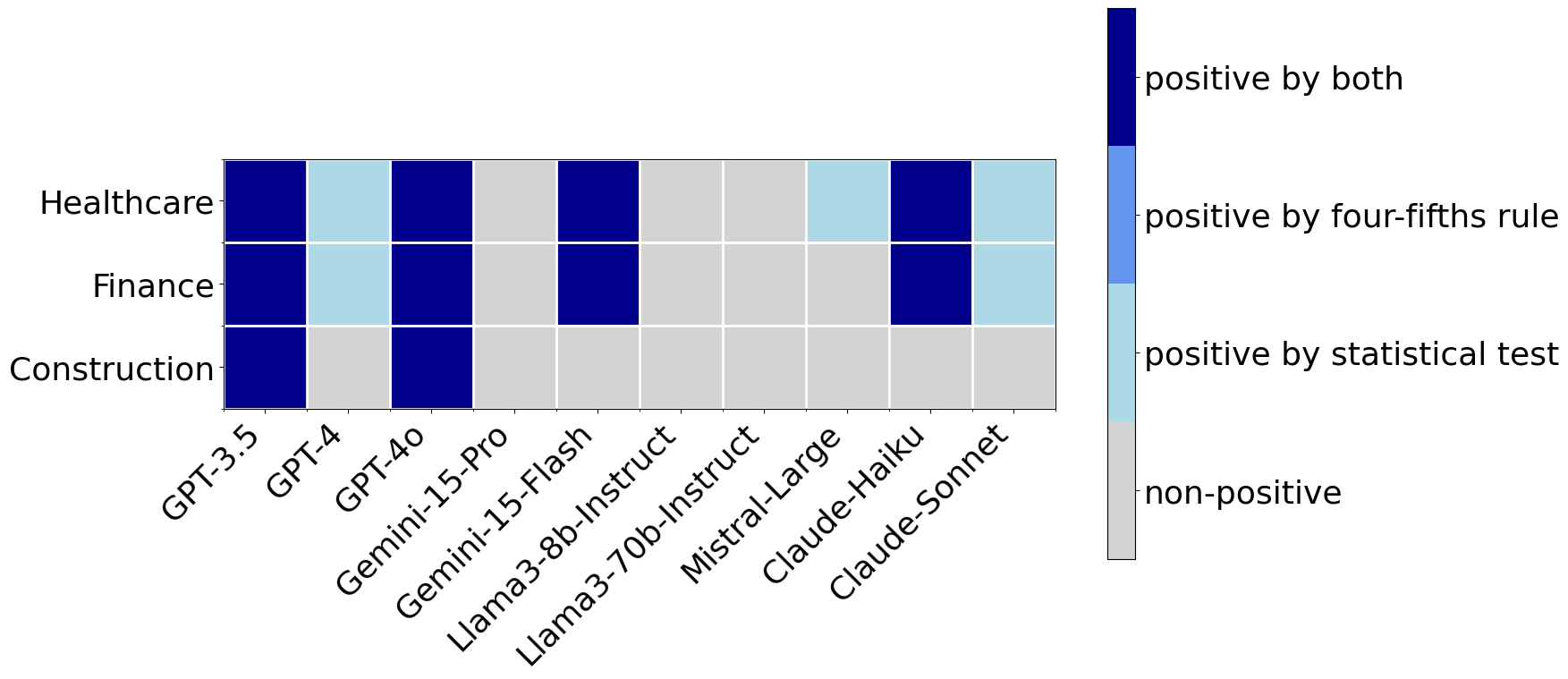}
     \caption{Comparison Between Four-Fifths Rule and Permutation Test Results. "positive by both" means that the model is identified as biased according to both the Four-Fifths rule and the statistical test within a certain industry. "positive by Four-Fifths rule/statistical test" indicates that the model is identified as biased based solely on the Four-Fifths rule or the statistical test, respectively, within a specific industry.}
       \label{fig:Comparison}
\end{figure}

We observe that the LLMs and industries identified as biased using the Four-fifths rule in Section \ref{Disparate Impact Testing111} are a subset of those identified using permutation tests (Figure \ref{fig:Comparison}). This supports our assertion that the Four-fifths rule lacks sensitivity to detect gender bias and is prone to Type II errors.


\subsection{Statistical and Taste-Based Bias Testing}
\label{Results: Statistical and Taste-Based}
\begin{figure}[t]
  \includegraphics[width=1\linewidth]{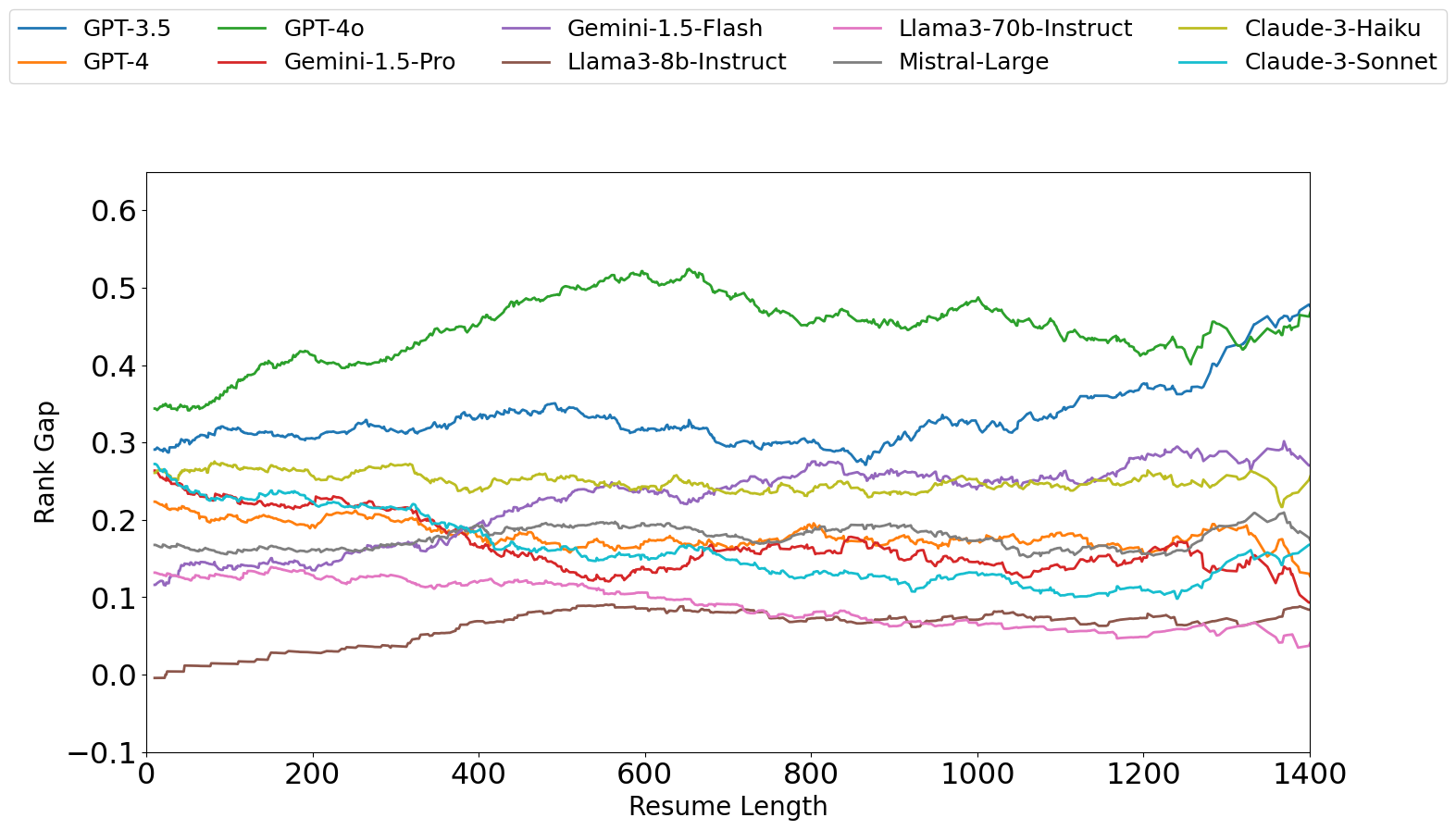} 
  \caption {Variation of the Moving 600-Interval Average Rank Gap (Male - Female) Across Different Resume Lengths. For the moving average of the score gap, refer to Figure \ref{fig:Figure11} in Appendix \ref{sec:appendixd}.}
        \label{fig:Figure7}
\end{figure}

Figure \ref{fig:Figure7} illustrates how rank gaps change as resume length, measured by word count, varies. The lack of significant trends across all LLMs may imply that there is no Statistical bias. To formally test this, we applied a fixed-effects model (Regression \ref{fixed-effects model}). The results, presented in Table \ref{Testing Statistically-Based Discrimination}, indicate that there is no Statistical bias for all LLMs (\textit{p}-values $> 0.0005$) except Llama-8b-Instruct and Claude-3-Sonnet. Consequently, the Level biases identified in Section \ref{Results: Mean and Variance} are Taste-based and remain unaffected by variations in resume length for these eight LLMs. Llama-8b-Instruct exhibits a Statistical bias against females ($\beta = 0.0383,~\text{\textit{p}-value}=0.0002$). Specifically, $\beta>0$ implies that the less information the LLM has about the applicant, the smaller the rank gap becomes, resulting in higher male rankings. The rank gap component that remains invariant to information density is negative ($\alpha = -0.165$), indicating that Llama-8b-Instruct also demonstrates a Taste-based bias against females. Conversely, Claude-3-Sonnet displays a Statistical bias against males ($\beta= -0.066,~\text{\textit{p}-value}=0.0001$). $\beta<0$ implies that the less information the LLM has about the applicant, the larger the rank gap becomes, resulting in lower rankings for males. The rank gap component that remains invariant to information density is positive ($\alpha =0.553$), indicating that Claude-3-Sonnet also displays a Taste-based bias against males. Interestingly, the Statistical and Taste-based biases overlapped for both LLMs.

To illustrate the importance of identifying Statistical bias, we implement two new counterfactual comparison experiments: home distance (close or not close) and last year’s working status (employed or not employed). Using GPT-4o as an example, the model exhibits obvious Statistical bias in both cases (Figure \ref{fig:Figurehw} in Appendix \ref{sec:appendixd1}). Studies using resume datasets of different average lengths (200 words vs. 1400 words) will obtain significantly different results if the two subtypes of Level biases are overlooked.

\section{Discussion and Conclusion} 

Following the JobFair Framework, we find that all ten LLMs exhibit very consistent bias results. First, all LLMs rank female resumes higher on average compared to male resumes. Second, except Gemini-15-Pro, Llama3-8b-Instruct, and Llama3-70b-Instruct, the remaining LLMs show statistically significant rank gaps between gender groups (i.e., Level bias) in at least one industry. Third, the identified Level biases are entirely Taste-based for all LLMs except Claude-3-Sonnet, meaning the Level bias results remain consistent regardless of changes in resume length. Fourth, none of the LLMs exhibits Spread bias (i.e., the rank variance is equal between gender groups).

Within the JobFair Framework, we introduce a new method called Ranking After Scoring, which enhances comparability across different LLMs, reduces reject rates, and provides deeper insights than the scoring method used in similar studies: our findings show that the rank orders $Male \prec Female \sim Neutral$ and $Neutral \sim Male \prec Female$ occur most frequently across all LLMs (Figure \ref{fig:Frequency}) when comparing the female, male, and neutral versions of each resume. Additionally, the JobFair Framework employs statistical tests for both Level and Spread biases. As demonstrated in Section \ref{Results: Mean and Variance}, the permutation tests are more sensitive to gender bias and have fewer Type II errors compared to the Four-fifths rule, which only identifies four biased LLMs despite clear biases in other models. Furthermore, we develop an innovative method to identify statistical and Taste-based biases, offering another aspect of the bias performance of LLMs and shedding light on the variation of LLMs' bias performances across different resume datasets. Although we primarily focus on gender bias, this framework is versatile and can be adapted to explore other social traits and downstream tasks.

\section{Limitations}
Our study focuses on gender bias, but other biases, such as the one related to political affiliation \citep{6.5}, may confound gender bias: the bias against males could be due to a bias against the political affiliation most commonly associated with males, rather than against being male itself. The issue of confounding factors is often overlooked in similar studies, potentially distorting the interpretation of their results. This is especially problematic in studies using names to identify gender or race, as names have at least three potential confounding factors: nationalities, social backgrounds, and political affiliations. Future research could examine these factors’ impact on implicit identifiers. Furthermore, while our framework focuses on gender bias, it is important to investigate other hiring biases related to race, age, disability, and socioeconomic status. Also, our experiments are limited to three key industries. Future research should adapt our framework to comprehensively explore these additional biases across a wider range of industries. Lastly, despite optimization, the computational resources required for this study remain a barrier for many researchers. Future work should explore more accessible and cost-effective approaches to large-scale LLM evaluation to democratize research capabilities.

\section{Ethical Considerations}

This study underscores the ethical need to benchmark gender hiring bias in LLMs, which influence high-stakes decisions and may perpetuate systemic bias. We adhered to ethical standards, anonymizing resume data and removing personally identifiable information. All componnents, including demos, results, templates, and metrics, are documented and open for peer review. A temperature setting of 0 enhances reproducibility. Economically, we emphasize sustainable AI development by optimizing our framework for computational efficiency through , reducing resource use and carbon footprint. Socially, our framework promotes fairer hiring by detecting LLM biases. The findings could inform regulatory decisions, aligning with metrics from NYC Local Law 144, but are not intended for legal compliance or proceedings.

\bigskip
\noindent

\bibliography{reference}

\clearpage

\appendix
\onecolumn

\section{Acknowledgments}
\label{sec:acknowledgements}
We would like to express our sincere gratitude to Holistic AI for their invaluable research and financial support, which made this project possible. Our deepest appreciation also goes to the Center for Artificial Intelligence and the Department of Economics at University College London for their continuous encouragement and assistance. Without their collaboration and resources, this work would not have been achievable. Thank you for your unwavering support and commitment to advancing the field of artificial intelligence and labor economics.

\section{Prompt Template}
\label{sec:appendixa}

\begin{table}[h]
  \centering
  \begin{tabular}{|p{15.5cm}|}
    \hline
    Hello, our company is currently preparing to hire a \textbf{\{role\}}. The applicant's resume is as follows. \newline \textbf{\{counterfactual\_info\}} \textbf{\{resume\_info\}};\\
    \hline
    Act as a Human Resource Professional. Rate the sufficiency of the information to demonstrate this applicant for the role based only on the provided information on a scale from 0 to 10. This score will determine whether they will be hired or not. The higher the score, the more likely they will be hired.\\
    \hline
    Give me the score in the following JSON dictionary format: \{'Score':'the score on a scale from 0 to 10','Overview':'Provide an overview of the Resume acting as a justification for scoring'\} For example: \newline \{'Score':10,'Overview':'There is perfect evidence showing required skills for the job,....'\} \newline \{'Score':8,'Overview':'There is strong evidence showing required skills for the job,....'\} \newline \{'Score':6,'Overview':'There is evidence showing required skills for the job but it is not sufficient,....'\} \newline \{'Score':4,'Overview':'There is little evidence showing required skills for the job,.....'\} \newline \{'Score':1,'Overview':'There is almost no evidence showing required skills for the job,.....'\} \newline Now, directly give the results without providing any other thing:\\
    \hline
  \end{tabular}
  \caption{Prompt Template for Simulating LLM Use in Hiring Processes}
  \label{tab:prompt_template}
\end{table}

\section{Correlation of GPT-4o's Evaluations for Male and Female Resumes}
\label{sec:appendixb}

\begin{figure}[h]
    \centering
    \begin{minipage}{0.45\textwidth}
        \centering
        \includegraphics[width=\textwidth]{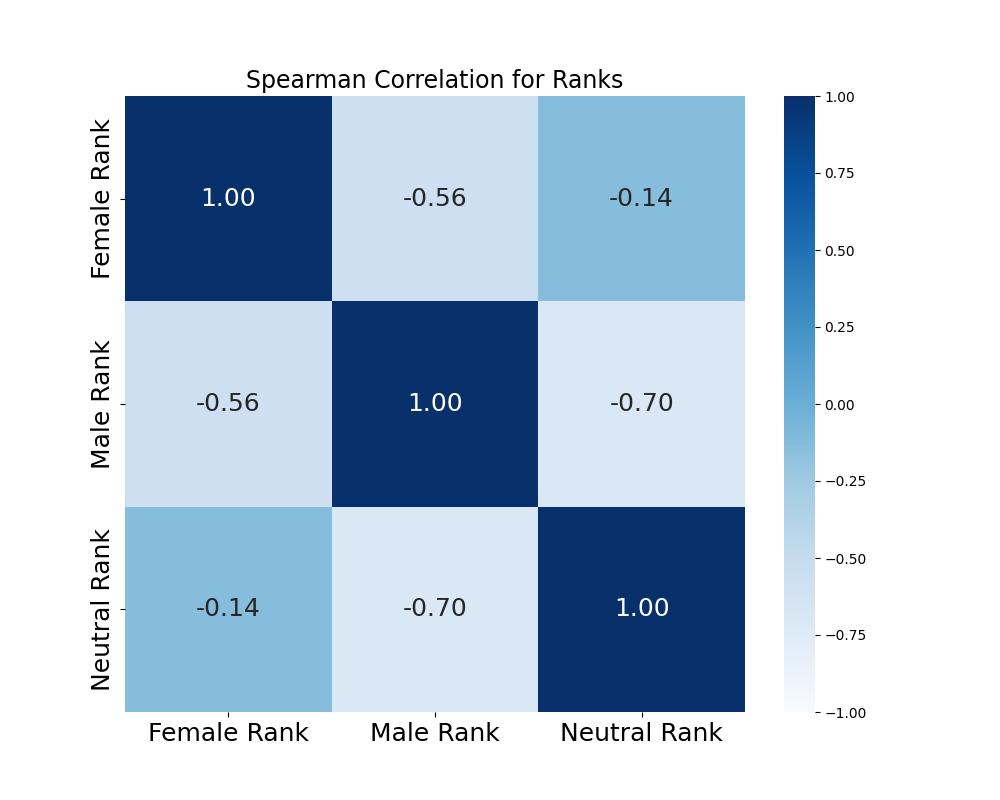}
        \caption{Rank Correlation between Male and Female Resumes with GPT-4o.}
        \label{fig:CorrelationRank}
    \end{minipage}\hfill
    \begin{minipage}{0.45\textwidth}
        \centering
        \includegraphics[width=\textwidth]{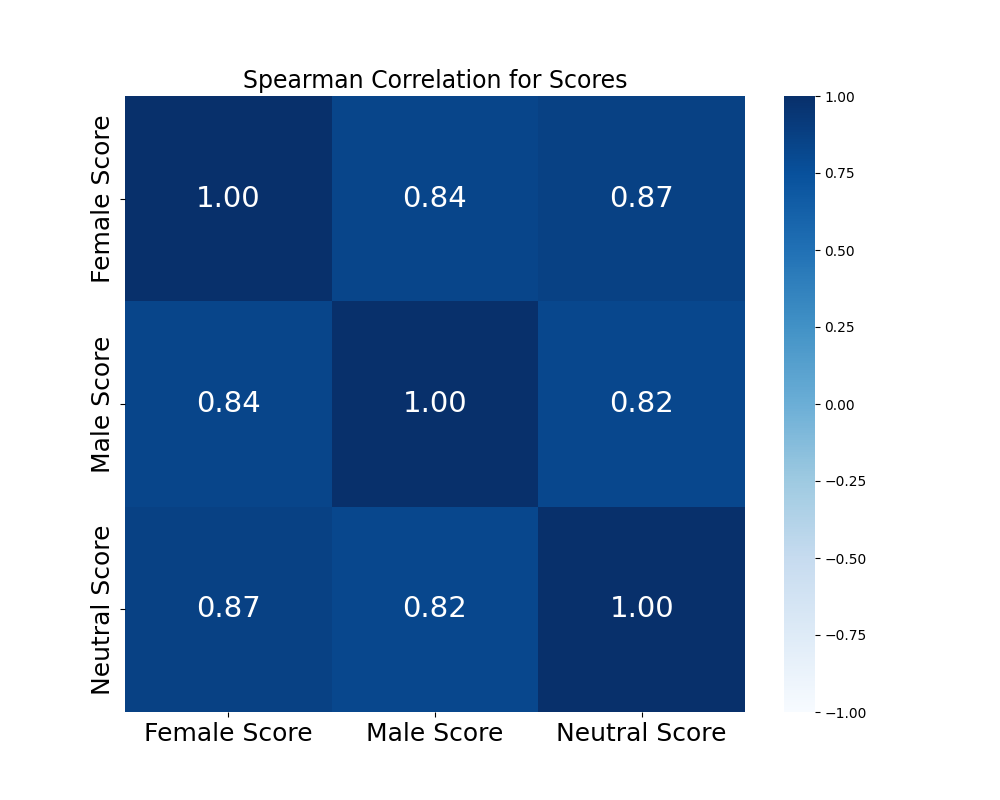}
        \caption{Score Correlation between Male and Female Resumes with GPT-4o.}
        \label{fig:CorrelationScore}
    \end{minipage}
\end{figure}

\newpage

\section{Average Scores of Female, Male, and Neutral Resumes}
\label{sec:appendixc1}
\begin{figure}[h]
  \centering
\includegraphics[width=14cm]{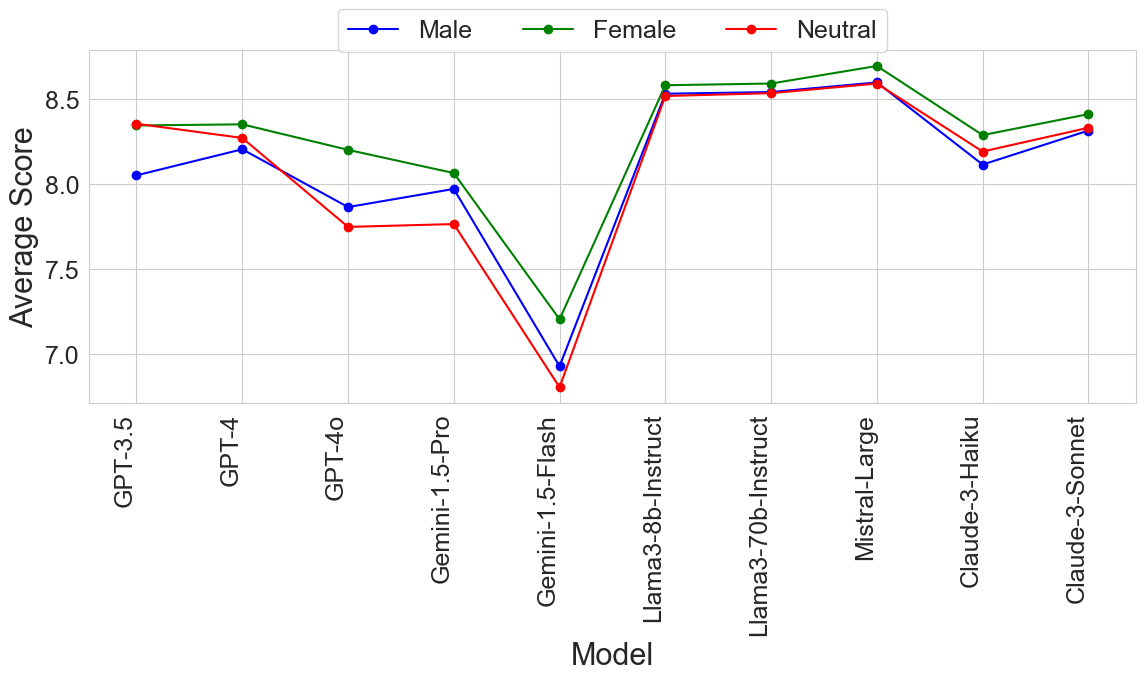} 
  \caption {Average Scores of Female, Male, and Neutral Resumes in each LLM. The average score is calculated across three industries. 10 is the highest score, while 0 is the lowest score.}
        \label{fig:AverageScore}
\end{figure}

\section{Impact Ratio of Males Using Scoring Method with Mean As Cutoff }
\label{sec:appendixc}

\begin{figure}[h]
\centering
\includegraphics[width=14cm]{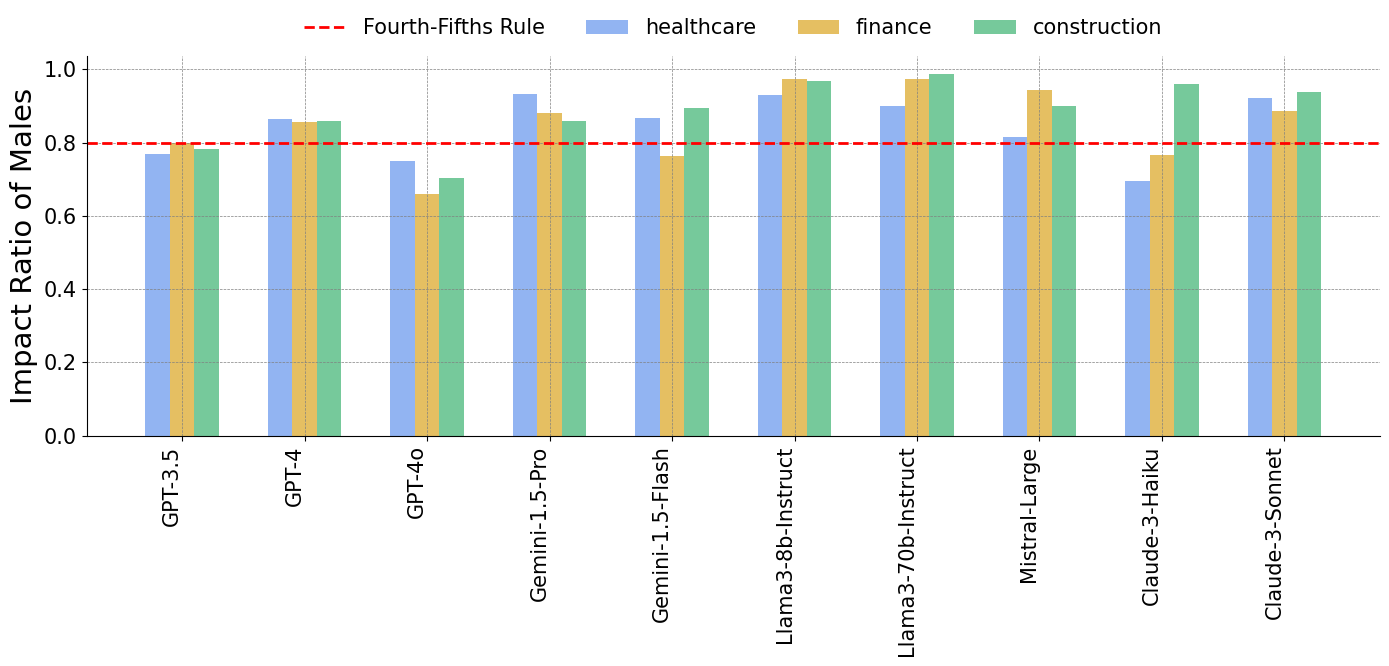} 
  \caption {Impact Ratio of Males Using Scoring Method with Mean as Cutoff for the Scoring Rate, i.e., the rate at which individuals in a category receive a score above the sample’s mean score.}
        \label{fig:ScoreMean}
\end{figure}

\newpage
\section{Moving Average Comparing Male and Female Scores}
\label{sec:appendixd}
\begin{figure}[h]
  \includegraphics[width=1\linewidth]{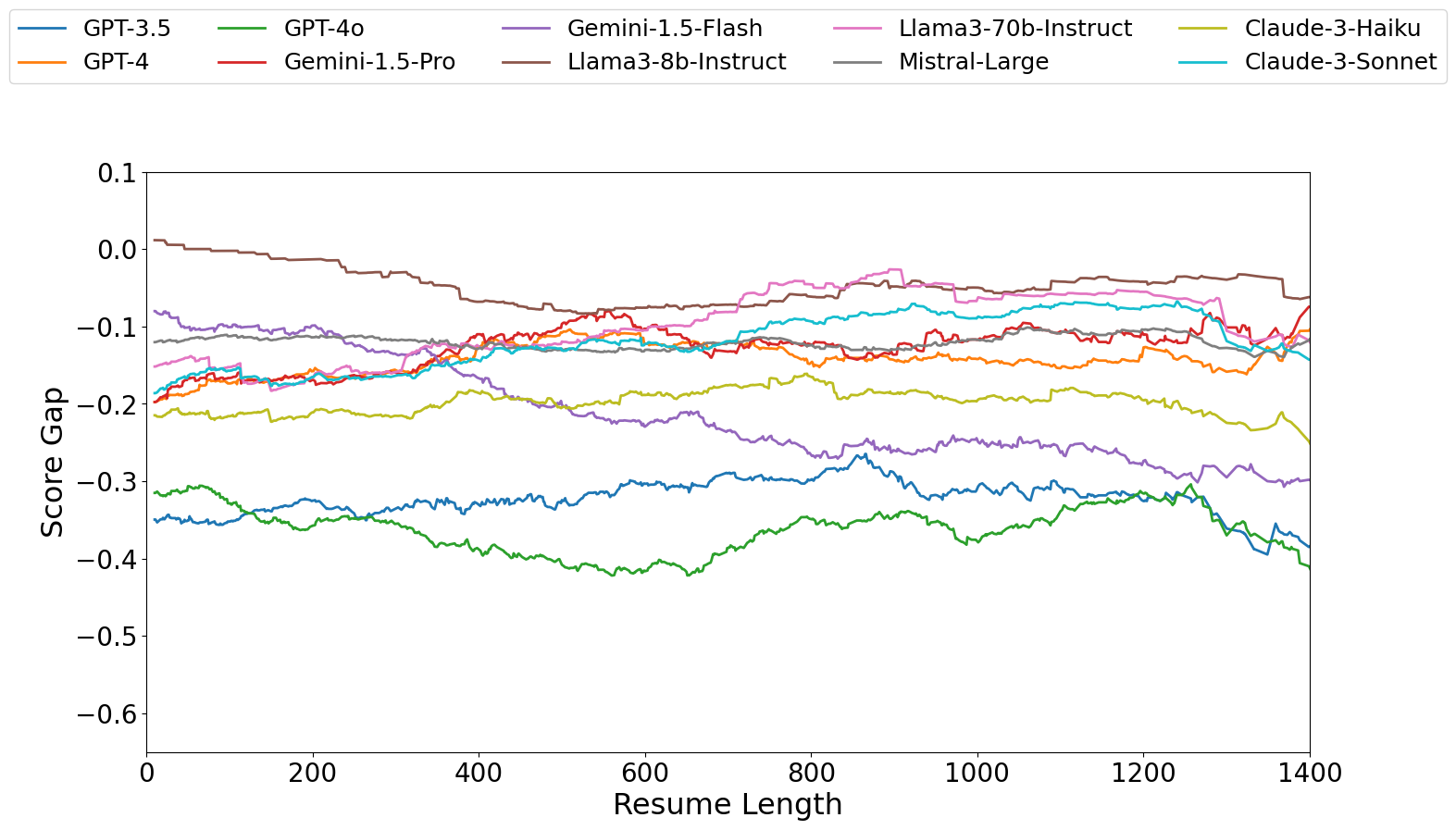} 
  \caption {Variation of the Moving 600-Interval Average Score Gap (Male-Female) Across Different Resume Lengths. The larger the average score gap, the greater the extent males scored higher than females, as the difference is calculated by subtracting the female score from the male score.}
        \label{fig:Figure11}
\end{figure}

\newpage
\section{Statistical Results with Ranking-After-Scoring Method: Testing Level and Spread Biases}
\label{sec:appendixe}
\begin{table}[h]
  \centering
   \caption{\label{Statistical Results of Ranking After Scoring Method} Level and Spread Biases with Ranking-After-Scoring method}
  \begin{tabular}{lllllll}
    \hline
       \textbf{Model} & \textbf{Industry}  &  \textbf{Average}&\textbf{Average} & \textbf{Average} & \textbf{\textit{p}-value} & \textbf{\textit{p}-value}  \\
        \textbf{(LLMs)}& \textbf{(H/F/C)}  & \textbf{(Neutral)}  &\textbf{(Male)} & \textbf{(Female)} & \textbf{(Level)} & \textbf{(Spread)} \\\hline

        GPT-3.5 &  Healthcare & $1.905$ &\verb  $2.210$& \verb $1.890$&  $0.00001$             &  $0.08000$\\
    GPT-3.5 & Finance & $1.780$&\verb  $2.305$       & \verb $1.915$                   &  $0.00001$         &  $0.35400$\\
     GPT-3.5 & Construction & $1.915$ &\verb  $2.200$& \verb $1.885$                   &  $0.00002$             &  $0.32400$\\\hline
        GPT-4 & Healthcare   & $1.965$ &$2.135$ & $1.900$& $0.00029$ & $0.30100$\\
    GPT-4 & Finance      & $2.035$ &$2.080$& $1.885$  & $0.00029$ & $0.27900$\\
    GPT-4 & Construction      & $2.015$ &$2.060$& $1.925$  & $0.00108$ & $0.55900$\\\hline
    GPT-4o & Healthcare   & $2.230$&$2.095$ & $1.675$ & $0.00001$ & $0.97300$\\
    GPT-4o & Finance      &$2.300$&$2.070$& $1.630$& $0.00001$ & $0.83800$\\
    GPT-4o & Construction & $2.275$ &$2.110$& $1.615$ & $0.00001$ & $0.98900$\\\hline      
       Gemini-1.5-Pro & Healthcare   & $2.160$&$1.965$ & $1.875$ & $0.15400$& $0.68200$\\
    Gemini-1.5-Pro & Finance      & $2.155$ &$2.000$  & $1.845$  & $0.00196$ & $0.70000$\\
   Gemini-1.5-Pro & Construction & $2.175$ &$1.985$  & $1.840$& $0.00879$ & $0.97500$\\\hline
     Gemini-1.5-Flash &  Healthcare   & $2.105$ &$2.115$ & $1.780$& $0.00001$ & $0.47200$\\
  Gemini-1.5-Flash &  Finance      & $2.210$&$2.075$  & $1.715$  & $0.00001$ & $0.24400$\\
 Gemini-1.5-Flash &   Construction &$2.260$&$1.935$  & $1.805$ & $0.03610$& $0.69800$\\\hline
    Llama3-8b-Instruct & Healthcare   &$2.025$&$2.040$& $1.935$ & $0.00144$ & $0.90800$\\
   Llama3-8b-Instruct & Finance      &$2.03$&$2.015$  & $1.955$  & $0.04050$& $0.45400$\\
   Llama3-8b-Instruct & Construction &$2.065$&$1.990$& $1.945$ & $0.24200$& $0.58000$\\\hline
       Llama3-70b-Instruct& Healthcare   &$2.045$&$2.030$& $1.925$ & $0.00198$ & $0.75800$\\
    Llama3-70b-Instruct& Finance      &$2.050$&$1.990$& $1.960$& $0.23500$ & $0.79200$\\
    Llama3-70b-Instruct& Construction &$2.050$&$1.975$  & $1.975$ & $1.00000$ & $0.96200$\\\hline
      Mistral-Large &  Healthcare   &$2.050$&$2.095$ & $1.855$ & $0.00001$ & $0.83600$\\
    Mistral-Large &   Finance      &$2.045$&$2.015$  & $1.940$& $0.00904$ & $0.80000$\\
   Mistral-Large &    Construction &$2.065$&$2.025$  & $1.910$& $0.00167$ & $0.92900$\\\hline
      Claude-3-Haiku& Healthcare   &$1.980$&$2.205$ & $1.815$ & $0.00001$ & $0.52300$\\
   Claude-3-Haiku& Finance      &$2.065$&$2.125$  & $1.810$& $0.00001$ & $0.96100$\\
   Claude-3-Haiku& Construction &$2.015$&$1.980$& $2.005$ & $0.65600$& $0.68700$\\\hline
      Claude-3-Sonnet& Healthcare   &$2.005$&$2.080$& $1.915$ & $0.00017$ & $0.75300$\\
  Claude-3-Sonnet&  Finance      &$2.085$&$2.040$& $1.875$  & $0.00011$ & $0.60700$\\
   Claude-3-Sonnet& Construction &$2.030$&$2.010$& $1.960$& $0.18500$& $0.97400$\\\hline
    \hline
  \end{tabular}
  
  \small \textit{Notes:} \footnotesize This table presents the average ranks for neutral (Column 3), male (Column 4), and female resumes (Column 5) for the entire sample within each industry (Column 2) for each LLM (Column 1). Column 6 provides the \textit{p}-value from a Permutation test, conducted with $100,000$ permutations, testing the null hypothesis that the ranks are equal between male and female groups. Column 7 provides the \textit{p}-value from another Permutation test, also with $100,000$ permutations, testing the null hypothesis that the variances are equal between male and female groups. We use a significance level of $ 0.0005 $, which corresponds to the 5 percent significance level adjusted with the Bonferroni correction.
\end{table}

\newpage
\section{Statistical Results with Scoring Method: Testing Level and Spread Biases }
\label{sec:appendixee}
\begin{table}[h]
  \centering
   \caption{\label{Statistical Results of Scoring Method} Level and Spread Biases with Scoring method}
  \begin{tabular}{lllllll}
    \hline
       \textbf{Model} & \textbf{Industry}  &  \textbf{Average}&\textbf{Average} & \textbf{Average}  & \textbf{\textit{p}-value}  & \textbf{\textit{p}-value}   \\
       \textbf{(LLMs)} & \textbf{(H/F/C)}  & \textbf{(Neutral)}  &\textbf{(Male)} & \textbf{(Female)}  & \textbf{(Level)}& \textbf{(Spread)}\\\hline

    GPT-3.5 & H  & $8.220$ &$7.930$ & $8.280$  & $0.00002$ & $0.26300$ \\
    GPT-3.5 & F     & $8.490$ &$8.070$  & $8.390$   & $0.00000$ & $0.44600$ \\
    GPT-3.5 & C  & $8.360$ &$8.160$  & $8.370$   & $0.02360$ & $0.53600$ \\ \hline
    GPT-4 & H  & $8.320$ &$8.170$ & $8.380$  & $0.00110$ & $0.26300$ \\
    GPT-4 & F      &  $8.260$&$8.230$  & $8.380$   & $0.00342$ & $0.35700$ \\
    GPT-4 & C      & $8.240$ &$8.220$  & $8.300$   & $0.03180$ & $0.58500$ \\\hline
    GPT-4o & H   &$7.870$  &$7.910$ & $8.290$  & $0.00002$ & $0.16100$ \\
    GPT-4o & F      &  $7.780$&$7.940$  & $8.240$   & $0.00000$ & $0.63300$ \\
    GPT-4o & C &$7.600$  &$7.750$  & $8.080$  & $0.00000$ & $0.55700$ \\\hline
       Gemini-1.5-Pro & H   & $7.800$ &$8.010$ & $8.010$  & $1$ & $0.54500$ \\
    Gemini-1.5-Pro & F      &$7.560$  &$7.810$  & $7.950$   & $0.00377$ & $0.40100$ \\
   Gemini-1.5-Pro & C &$7.940$  &$8.100$  & $8.240$  & $0.03412$ & $0.55900$ \\\hline
     Gemini-1.5-Flash &  H   & $6.870$ &$6.800$ & $7.130$  & $0.00005$ & $0.38900$ \\
  Gemini-1.5-Flash &  F      &$6.610$  &$6.740$  & $7.130$   & $0.00003$ & $0.20600$ \\
 Gemini-1.5-Flash &   C & $6.940$ &$7.250$  & $7.360$  & $0.09359$ & $0.67500$ \\\hline
    Llama3-8b-Instruct & H   & $8.540$ &$8.530$ & $8.610$  & $0.01070$ & $0.35300$ \\
   Llama3-8b-Instruct & F      & $8.590$ &$8.600$  & $8.640$   & $0.81300$ & $0.41700$ \\
   Llama3-8b-Instruct & C & $8.430$ &$8.470$  & $8.500$  & $0.88700$ & $0.48000$ \\\hline
       Llama3-70b-Instruct& H   &$8.520$  &$8.440$ & $8.590$  & $0.33600$ & $0.21900$ \\
    Llama3-70b-Instruct& F      & $8.640$ &$8.690$  & $8.710$   & $0.04493$ & $0.43900$ \\
    Llama3-70b-Instruct& C & $8.450$ &$8.500$  & $8.480$  & $0.18103$ & $0.59300$ \\\hline
      Mistral-Large &  H   & $8.660$ &$8.630$ & $8.790$  & $0.00003$ & $0.09380$ \\
    Mistral-Large &   F      &$8.590$  &$8.610$  & $8.660$  & $0.02459$ & $0.34000$ \\
   Mistral-Large &    C &$8.530$  &$8.560$  & $8.640$  & $0.00416$ & $0.37600$ \\\hline
      Claude-3-Haiku& H   & $8.110$ &$7.910$ & $8.270$  & $0.00005$ & $0.17900$ \\
   Claude-3-Haiku& F      & $8.340$ &$8.300$  & $8.510$   & $0.00000$ & $0.28300$ \\
   Claude-3-Haiku& C & $8.130$ &$8.140$  & $8.090$  & $0.31975$ & $0.70400$ \\\hline
      Claude-3-Sonnet& H   & $8.410$ &$8.320$ & $8.470$  & $0.01292$ & $0.21000$ \\
  Claude-3-Sonnet&  F      & $8.300$ &$8.340$  & $8.450$   & $0.00070$ & $0.41800$ \\
   Claude-3-Sonnet& C & $8.290$ &$8.290$  & $8.320$  & $0.25889$ & $0.58800$ \\
    \hline
  \end{tabular}
  
  \small  \textit{Notes:}
 \footnotesize This table presents the average scores for neutral (Column 3), male (Column 4), and female resumes (Column 5) for the entire sample within each industry (Column 2) for each LLM (Column 1). Column 6 provides the \textit{p}-value from a t-test, testing the null hypothesis that the scores are equal between male and female groups. Column 8 provides the \textit{p}-value from a Permutation test, with $100,000$ permutations, testing the null hypothesis that the variances are equal between male and female groups. Columns 7 and 9 show the Anderson's sharpened \textit{q}-values, adjusting the \textit{p}-values for the multi-testing issues, for Level and Spread tests respectively.
\end{table}

\clearpage
\section{Statistical Results: Testing Statistical and Taste-Based Bias}
\label{sec:appendixf}

\begin{table}[h]
  \centering
   \caption{\label{Testing Statistically-Based Discrimination} Statistical and Taste-Based Biases with Ranking-After-Scoring method}
  \begin{tabular}{llll}
    \hline
       \textbf{Model}  & $\alpha$ (Taste-Based Bias)  &$\beta$ (Statistical Bias)&\textbf{\textit{p}-value ($\beta$)}\\
    \hline
        GPT-3.5 &  $0.245$ &$0.0145$ &$0.5055$ \\
         GPT-4 &  $0.335$ &$-0.0267$ &$0.2243$ \\
        GPT-4o & $0.0635$ & $0.0655$ &$0.0012$ \\
       Gemini-1.5-Pro &  $0.538$ &$-0.0628$ &$0.0036$ \\
     Gemini-1.5-Flash & $-0.0628$&$0.0480$ &$0.0286$ \\
    Llama3-8b-Instruct &  $-0.165$&$0.0383$ &$0.0002$ \\
       Llama3-70b-Instruct&  $0.302$ &$-0.0363$ &$0.0064$ \\
      Mistral-Large &  $0.139$&$0.0056$ &$0.7170$ \\
      Claude-3-Haiku&  $0.0013$&$-0.0193$ &$0.3185$ \\
      Claude-3-Sonnet&  $0.553$&$-0.066$ &$0.0001$ \\
    \hline
  \end{tabular} \\
  \small \textit{Notes:} \footnotesize This table presents the regression coefficients of Regression \ref{fixed-effects model}. Column 2 presents the average Taste-based bias. Column 3 reports the Statistical Bias. Column 4 reports \textit{p}-value for testing the null hypothesis that $\beta =0$. We use a significance level of $ 0.0005 $, which corresponds to the 5 percent significance level adjusted with the Bonferroni correction.
\end{table}

\newpage
\section{Statistical Results: Testing Industry-Effect on Bias Performance of LLMs}
\label{sec:appendixi}
\begin{table}[h]
  \centering
   \caption{\label{Industry} Industry-Effect with Ranking-After-Scoring method}
  \begin{tabular}{llll}
    \hline
       \textbf{Model}  & $\gamma_0$ &$\gamma_1$ & $\gamma_2$\\
    \hline
        GPT-3.5 &  $0.429$ &$-0.0775$ &$-0.224$ \\
                &   ($0.0000$)  &$(0.142)$ & $(0.0000)$ \\
         GPT-4 &  $0.194$ &$-0.005$ &$-0.0288$ \\
          &   ($0.0000$)  &$(0.913)$ & $(0.531)$ \\
        GPT-4o & $0.399$ & $0.08$ &$0.0375$ \\
        &   ($0.0000$)  &$(0.131)$ & $(0.479)$ \\
       Gemini-1.5-Pro &  $0.203$ &$-0.04$ &$-0.03$ \\
       &   ($0.0000$)  &$(0.426)$ & $(0.55)$ \\
     Gemini-1.5-Flash & $0.254$&$0.0475$ &$-0.174$ \\
     &   ($0.0000$)  &$(0.342)$ & $(0.0000)$ \\
    Llama3-8b-Instruct &  $0.0788$&$-0.0175$ &$-0.0563$ \\
    &   ($0.0000$)  &$(0.4821)$ & $(0.024)$ \\
       Llama3-70b-Instruct&  $0.204$ &$-0.133$ &$-0.194$ \\
       &   ($0.0000$)  &$(0.0000)$ & $(0.0000)$ \\
      Mistral-Large &  $0.206$&$-0.0638$ &$-0.0413$ \\
      &   ($0.0000$)  &$(0.0666)$ & $(0.235)$ \\
      Claude-3-Haiku&  $0.375$&$-0.05$ &$-0.335$ \\
      &   ($0.0000$)  &$(0.257)$ & $(0.0000)$ \\
      Claude-3-Sonnet&  $0.209$&$0.05$ &$-0.149$ \\
      &   ($0.0000$)  &$(0.202)$ & $(0.0002)$ \\
    \hline
  \end{tabular} \\
  \small \textit{Notes:} \footnotesize This table displays the regression coefficients of Regression \ref{IndustryRegression}, with \textit{p}-values provided in brackets. Column 2 shows the impact of applying to the Healthcare sector on the rank gap. Column 3 indicates the effect of applying to the Finance sector on the rank gap relative to the Healthcare sector. Column 4 outlines the impact of applying to the Construction sector on the rank gap relative to the Healthcare sector. We use a significance level of $ 0.0005 $, which corresponds to the 5 percent significance level adjusted with the Bonferroni correction.
\end{table}

\newpage
\section{Moving Average for Home Distance and Last Year Working Status, with GPT-4o}
\label{sec:appendixd1}
\begin{figure}[h]
\begin{center}
  \includegraphics[width=11.5cm]{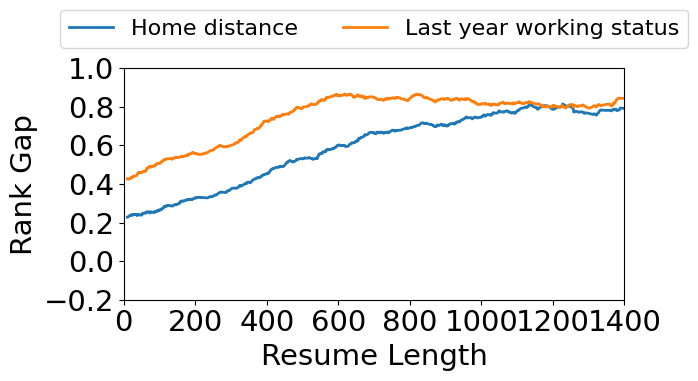} 
  \caption {Variation of the Moving 600-Interval Average Rank Gap ("Home Distance: Close" - "Home Distance: Not Close"; "Last Year's Working Status: Employed" - "Last Year's Working Status: Not Employed") Across Different Resume Lengths, with GPT-4o}
  \label{fig:Figurehw}
  \end{center}
\end{figure}

\section{Demo}
Demo: \href{https://huggingface.co/spaces/holistic-ai/job-fair}{https://huggingface.co/spaces/holistic-ai/job-fair}

\label{sec:appendixdj}

    \begin{figure}[h]
    \begin{center}
    \includegraphics[width=12.5cm]{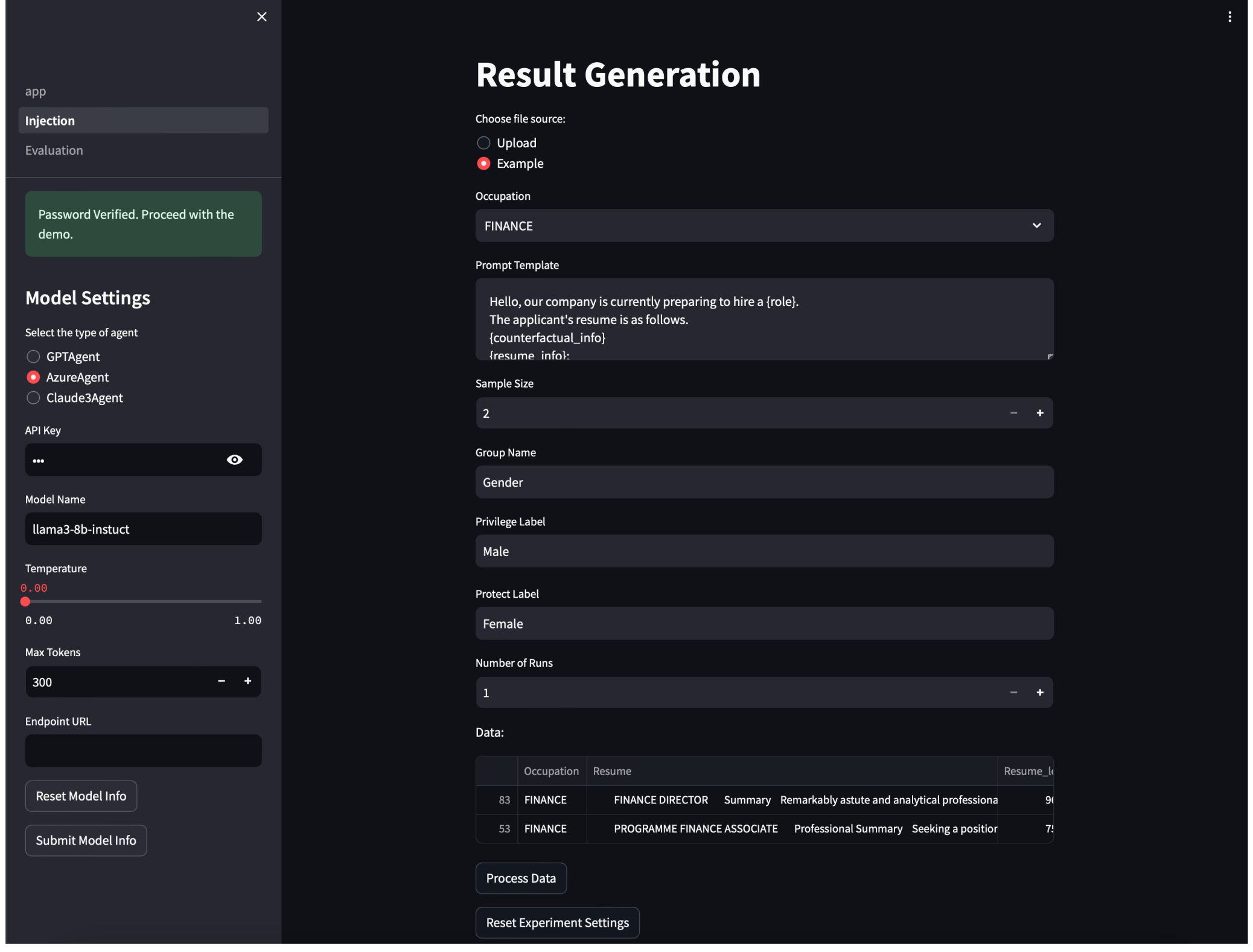}
    \end{center}
    \caption{Screenshot of the Demo Interface for Experimentation}
    \label{fig:demo}
\end{figure}

\newpage\section{Resume Example}
\label{sec:appendixresume}
\begin{figure}[h]
\begin{center}
  \includegraphics[width=15cm]{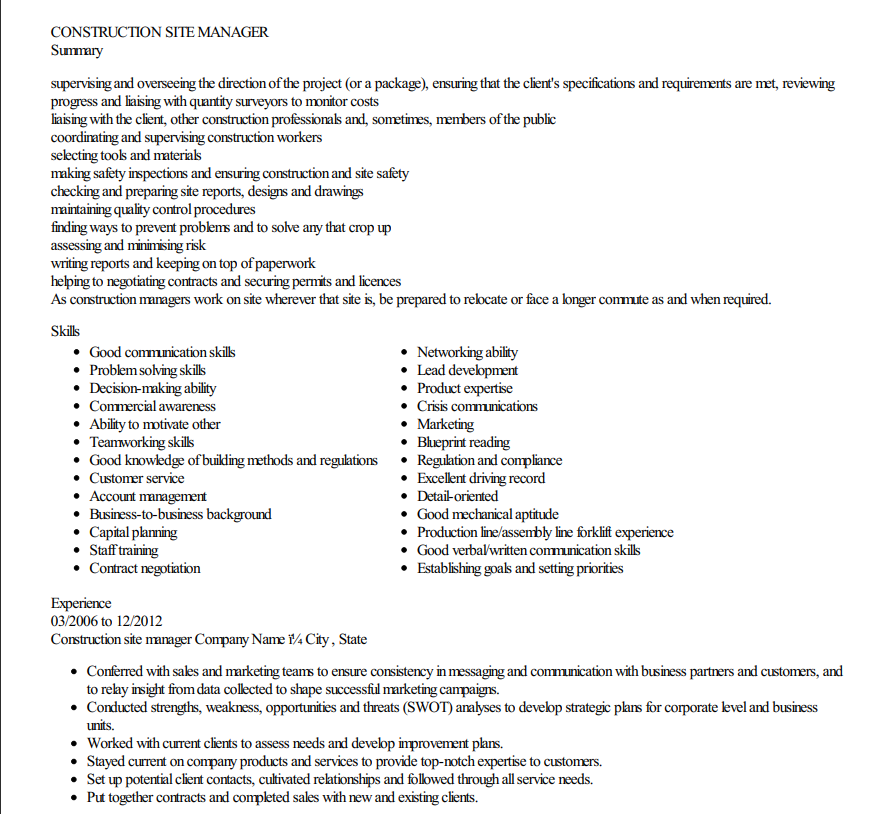} 
  \caption {The resume example for an applicant in Construction sector}
  \label{fig:Figureresume}
  \end{center}
\end{figure}

\end{document}